\documentclass[10pt, conference, compsocconf]{IEEEtran}

\usepackage{cite}

\ifCLASSINFOpdf
  \usepackage[pdftex]{graphicx}
      \DeclareGraphicsExtensions{.pdf,.jpeg,.png}
\else
        \usepackage[dvips]{graphicx}
      \DeclareGraphicsExtensions{.pdf,.jpeg,.png}
\fi

\usepackage{multirow}
\usepackage{amssymb}
\usepackage[cmex10]{amsmath}

\usepackage{algorithmic}

\usepackage{caption}
\usepackage[font=footnotesize]{subfig}

\usepackage{color}

\newcommand{\comment}[1]{}

\newcommand{\eg}{{\em e.g.}}
\newcommand{\ie}{{\em i.e.}}

\newcommand{\etal}{{\em et al.}}

\newcommand{\bs}{\boldsymbol}

\newcommand{\anthroParams}{{\bs\ell}}
\newcommand{\bshapeParams}{{\bs\beta}}
\newcommand{\meshParams}{{\Theta}}
\newcommand{\poseQ}{{\mathbf{q}}}
\newcommand{\meshVert}{{\mathbf{p}}}
\newcommand{\bonesMat}{{\mathbf{B}_\anthroParams}}
\newcommand{\bodyMat}{{\mathbf{B}_\bshapeParams}}
\newcommand{\meshModel}{{\mathcal{M}}}
\newcommand{\dataPt}{{\mathbf{d}}}
\newcommand{\geoDist}{d}
\newcommand{\geoExtr}{{\mathbf{g}}}

\begin{document}

\title{Walking on Thin Air: Environment-Free Physics-based Markerless Motion
Capture}

\author{\IEEEauthorblockN{Micha Livne\IEEEauthorrefmark{1}, \ \
Leonid Sigal\IEEEauthorrefmark{2}, \ \
Marcus A.\ Brubaker\IEEEauthorrefmark{3} \ \
and  \ \
David J.\ Fleet\IEEEauthorrefmark{1}  \hspace*{2.5cm} .
}
\and
\IEEEauthorblockA{\IEEEauthorrefmark{1}Department of Computer Science\\
University of Toronto\\
Toronto, Canada\\
\{mlivne, fleet\}@cs.toronto.edu
}
\and
\IEEEauthorblockA{\IEEEauthorrefmark{2}Department of Computer Science\\
University of British Columbia\\
Vancouver, Canada\\
lsigal@cs.ubc.ca}
\and
\IEEEauthorblockA{\IEEEauthorrefmark{3}Lassonde School of Engineering\\
York University\\
Toronto, Canada\\
mab@eecs.yorku.ca}
}

\maketitle

\begin{abstract}
We propose a generative approach to physics-based motion capture.
Unlike prior attempts to incorporate physics into tracking that assume
the subject and scene geometry are calibrated and known a priori,
our approach is automatic and online.  This distinction is important
since calibration of the environment is often difficult, especially for
motions with props, uneven surfaces, or outdoor scenes. The use of physics
in this context provides a natural framework to reason about contact and
the plausibility of recovered motions. We propose a fast
data-driven parametric body model, based on linear-blend skinning, which
decouples deformations due to pose, anthropometrics and body shape.
Pose (and shape) parameters are estimated using robust ICP optimization
with physics-based dynamic priors that incorporate contact.
Contact is estimated from torque trajectories and predictions of which
contact points were active.  To our knowledge, this
is the first approach to take physics into account without
explicit {\em a priori} knowledge of the environment or body dimensions.
We demonstrate effective tracking from a noisy single depth camera, improving
on state-of-the-art results quantitatively and producing better qualitative
results, reducing visual artifacts like foot-skate and jitter.
\end{abstract}


\IEEEpeerreviewmaketitle

\section{Introduction}

\begin{figure*}[ht]
\centering
\includegraphics[width=1.0\textwidth]{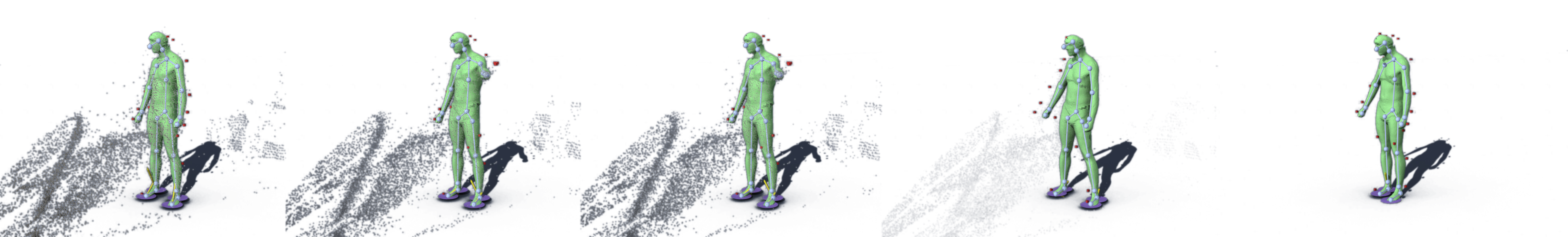}
\caption{Illustration of results produced by our physics-based tracker.}
\end{figure*}

Markerless motion capture methods enable reconstruction of detailed motion and
dynamic geometry of the body (and sometimes garments) from multiple streams of
video \cite{AguiarSIGGRAPH08} or depth data \cite{WeiSIGGRAPH12,DBLP:journals/corr/HaquePLAYL16}.
Recent human tracking methods are able to handle video captured in the wild,
but still suffer from visually significant artifacts (jittering, feet/contact skating).
This issue is significant as people are sensitive
to such artifacts (\eg, foot-skate is perceptible  at levels less  than 21 mm \cite{Prazak2011}).

To address these challenges we propose a generative 3D human tracking
approach that takes physics-based prior knowledge into account when estimating
pose over time. The use of physics in this context is compelling as it provides a natural
framework to reason about contact and the plausibility of the recovered
motions online. Prior attempts to use physics for tracking assume that the subject and scene
geometry are known a priori and calibrated \cite{BrubakerIJCV10,VondrakPAMI2013},
that contact states are annotated by a user \cite{WeiChaiSIGGRAPH10}, or that optimization can be
performed off-line (\ie, in batch) \cite{Brubaker2009}.
In contrast, our approach is online, without manual input.
Beginning with the first frame, the subject and the contact state(s)
are estimated online during tracking, without a priori knowledge
of the environment. This is an important distinction, as calibration of the environment can be
difficult, especially when capturing motions with props, on uneven surfaces or
outdoors.

Our main contribution is the use of a physics-based prior without an explicit model of the environment.
To our knowledge, it is the first tracking approach
to incorporate physics without any explicit {\em a priori} knowledge of the environment
or body dimensions. We demonstrate that the approach is effective in tracking from a single
depth camera, improving on state-of-the-art results quantitatively and
qualitatively, greatly reducing visually unpleasant artifacts such as
foot-skate and jitter.

\section{Related Work}

{\bf 3D Human Tracking:}
Markerless motion capture, estimating the skeletal motion of a subject,
has a rich history in vision and graphics (for an extensive survey see
\cite{Moeslund2006}). Methods can be broken into two classes: model-based and regression-based
(or generative vs. discriminative). Regression-based methods estimate pose
directly by regressing pose from image feature descriptors
(\eg \cite{Ren2005,ShottonCVPR11,Wang2009,DBLP:journals/corr/HaquePLAYL16,zhou2016sparseness,Tekin2016}).
Model-based approaches exploit a generative model for the body and image,
and optimize for generative parameters that explain the image
observations (\eg \cite{Deutscher2004,Felzenszwalb:2005:PSO:1024426.1024429}).
The former is faster, but generally less accurate (unless the
problem domain is highly constrained). Model-based approaches
may be more accurate, but  tend to be slower as they require iterative or stochastic
optimization, and suffer from local optima.

{\bf Use of Physics in Tracking:}
Physics-based tracking has been proposed as a way to regularize
pose under the assumption that physics is a universal prior
that requires no assumptions about one's motion (given a physical model).
Early work dates back to \cite{Metaxas1993} and \cite{Wren1998},
however they focused on simple motions in absence of contact.
More recently Brubaker \etal~\cite{BrubakerIJCV10} proposed a
low-dimensional model of the lower-body to track walking
subjects from monocular video.
A more general data-driven  physics-based filter, applicable to variety of motions,
was proposed in \cite{VondrakPAMI2013}.
In \cite{VondrakSIGGRAPH2012} a controller-based approach is proposed where
a physics-based full body controller, instead of sequence of poses,
is estimated.  In all cases the body proportions and the
scene geometry were assumed to be known.
In \cite{WeiChaiSIGGRAPH10} a physics-based tracking approach is formulated
as a batch optimization problem with known contact points and ground geometry.
In \cite{Brubaker2009} the parameters of the planar ground model are
estimated from data, however, the method assumes a parametric structure
of ground geometry (a plane) and reasonably accurate 3D input, obtained
using a binocular system.
We build on this work with one notable distinction: we assume no knowledge
of ground geometry or subject proportions.
A notable distinction from prior work is \cite{RosenhahnDAGM08}
which attempted to encode ground constraints directly in the kinematics,
without a physics model.  Like other methods, however, it required
{\em a prior} knowledge of the ground plane.

{\bf Contact Estimation:}
We briefly note that contact estimation and
sampling has been used in other domains of graphics as well. One
example is hand manipulation \cite{Liu2012}, where a randomized search
over the hand-object contacts is proposed as the strategy for finding pose
of the hand manipulating an object over time.
Contact invariant optimization \cite{Mordatch2012} attempts to sidestep
the problem of explicit contact estimation by searching over the space
of contacts at the same time as behaviour of the character.
Such approaches, while interesting, require batch processing and long
compute times, making them inapplicable for real-time capable
full body tracking.

\section{Method}

\begin{figure*}[t] \centering
\includegraphics[width=1.0\textwidth,height=7cm]{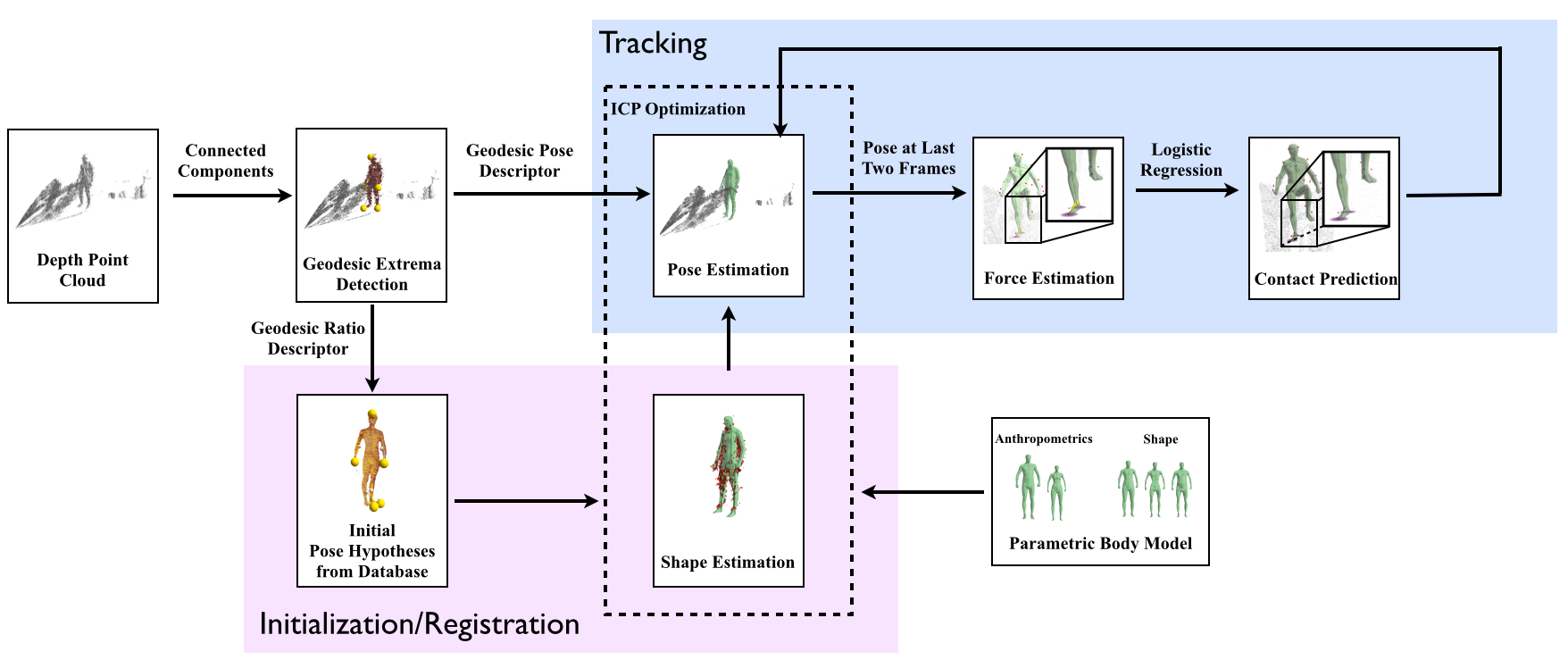}
\caption{An overview of the tracking pipeline. We extract
connected components and geodesic extrema from point cloud. Followed by
human detection and pose initialization. We initialize the pose with the
nearest-neighbour pose from a database, in geodesic descriptor space. Next, we
register body anthropometrics, shape, and pose. Tracking is performed by
updating pose with ICP optimization, while estimating contact state through
forces pattern.}
\label{fig:pipeline-overview}
\end{figure*}

Our tracking pipeline is depicted in Fig.\ \ref{fig:pipeline-overview}.
We describe in this section a fast body mesh model (Sec. \ref{subsec:fast data-driven parametric body model}),
discrete formulation of a physical engine and physics-based motion prior (Sec. \ref{subsubsec:discrete-mechanics}),
and a tracking framework that utilizes those in order to facilitate physics-based 3D human
tracking (Sec. \ref{subsec:registration-and-tracking}). We also describe a method to pre-process input point cloud
data that allows us to automatically initialize tracking, is fast, and simple to implement (Sec. \ref{subsec:pre-processing}).

\subsection{Fast Data-driven Parametric Body Model}
\label{subsec:fast data-driven parametric body model}

In what follows we exploit a new SCAPE-like model (see \cite{Anguelov2005}) for tracking.
With an explicit skeleton, anthropometrics (bone length) and body shape parameters,
our model is easy to manipulate and control.
The anthropometrics parameters offers direct control over deformations
due to bone lengths.  The body shape parameters allow for control over
the shape, independent of anthropometrics.
To the best of our knowledge, explicit control over anthropometrics and shape is
not straightforward with other existing body models.

\begin{figure}[t] \centering
\includegraphics[width=0.5\textwidth]{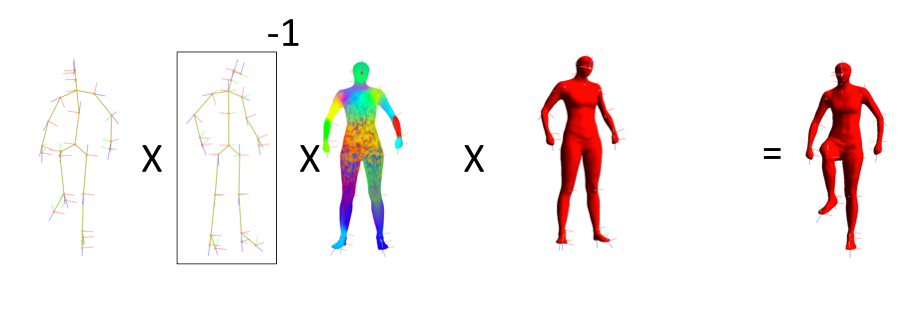}
\caption{Visualization of linear mesh blending. From left to right: target pose,
template pose, weights, template mesh, and target mesh. The final vertices
position vector $\meshVert(\poseQ)$ is calculated by weighing vertex position w.r.t.
transformations of a few joints applied to the template vertex position. }
\label{fig:linear-mesh-blending}
\end{figure}

\begin{figure}[t]
\centering
\includegraphics[width=0.3\textwidth,height=4cm]{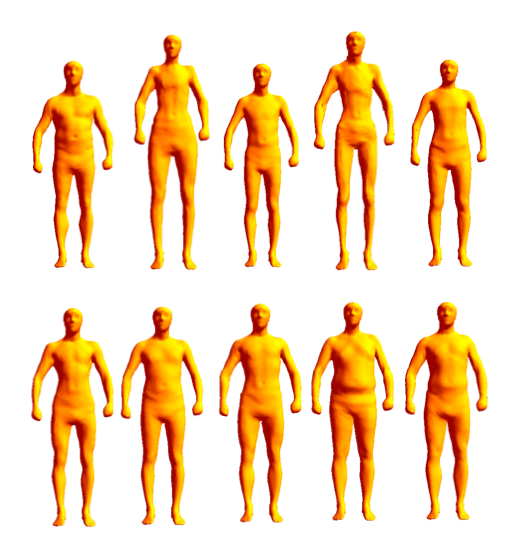}
\caption{A template is composed of anthropometrics $\anthroParams$, and body shape
$\bshapeParams$. Top row: anthropometrics variations with constant body shape.
Bottom row:
body shape variations with constant anthropometrics.}
\label{fig:model-template}
\end{figure}

The body is modelled as a 3D triangulated mesh and comprise 69 DOF and 26 body parts

\begin{equation}
\label{eq:model-overlook}
\meshModel \left( \meshParams \right) = \left\{ \meshVert \left(
  \meshParams; \bonesMat, \bodyMat \right), E \right\} ~
\end{equation}
where $\bonesMat$ and $\bodyMat$ are basis matrices, which
 capture variations in the mesh due to anthropometrics and body shape
respectively, and $\meshParams = ( \poseQ, \anthroParams,
\bshapeParams )$, where $\poseQ$ denotes articulated pose,
and $\anthroParams$ and $\bshapeParams$ denote coordinate vectors
within the two subspaces.
The $N$ mesh vertices of a canonical pose (called the template pose $\tilde{\poseQ}^s$)
for a given subject $s$ are given by a vector $\tilde{\meshVert}^s \in \mathbb{R}^{3N \times 1}$
\begin{equation}
\label{eq:model-template}
\tilde{\meshVert}^s ( \anthroParams^s, \bshapeParams^s )
= \bonesMat ( \anthroParams^s - \hat{\anthroParams} ) + \bodyMat \bshapeParams^s
\end{equation}
where $\hat{\anthroParams}$ denotes the mean anthropometrics within the
subspace. The anthropometrics basis $\bonesMat$ represents a linear mapping
from bone lengths (relative to the mean) to a base template mesh.
The basis $\bodyMat$ provides a linear mapping of body shape
coefficients into a deformation from a base template mesh.
We enforce orthogonality of the two subspaces during the basis learning stage.
We discuss how we learn the basis in the supplementary material.
The final mesh is calculated using Linear Mesh Blending (LMB)

\begin{equation}
\label{eq:linear-mesh-blending}
\meshVert_i^s( \meshParams) = \sum_{b \in \mathcal{B}_i}
w_{ib}
\mathbf{M}_b ( \anthroParams, \poseQ )
\mathbf{M}_b^{-1}(\anthroParams,\tilde{\poseQ}^s)
\tilde{\meshVert}_i^s (\anthroParams, \bshapeParams)
\end{equation}
where $\mathcal{B}_i$ is the set of bones (\ie, rigid body parts) that influence the position
of vertex $i$, $w_{ib}$ is the influence of bone $b$ on vertex $i$ (assumed to be constant for all poses),
and $\meshVert^s \in \mathbb{R}^{3N \times 1}$ is the final mesh. We trained our model on
the Hasler dataset \cite{Hasler2009}. Fig. \ref{fig:linear-mesh-blending} and Fig. \ref{fig:model-template}
depict our LMB model.

\subsection{Environment-Free Physics-Based Priors}
\label{subsec:physics-based-priors}

Priors are used in optimization to regularize the loss, pushing the optimal solution to a more desired manifold.
As such, we would like our priors to be as generic as possible in order to generalize well. Physics-based priors exploit
physical dynamics as an informative but general prior on motion, to help ensure that tracking yields
a plausible motion. To that end we formulate our model of articulated dynamics using
discrete mechanics \cite{Marsden2001}.  This has many desirable properties
such as direct mapping to discrete observations, conservation of energy, and
computational efficiency (see \cite{Johnson2009}).

{\bf Variational Integrator:}\label{subsubsec:discrete-mechanics} In the variational formulation of
Lagrangian mechanics, the motion of a system is described by a function known as a \emph{discrete Lagrangian},
$\mathcal{L}^d(\poseQ_{k-1}, \poseQ_k)$ where $\poseQ_k$ denotes the
generalized coordinates of the system (\eg, a stick figure) at time step $k$.
The discrete Lagrangian is an approximation of the continuous Lagrangian,
and is used in a discrete formulation of the principal of least action
to derive discrete mechanics (see \cite{Marsden2001} for more details).
The evolution of the system is then given by discrete Euler-Lagrange equations
\begin{equation}
\label{eq:discrete-euler-lagrange-equations}
\underbrace{
D_1 \mathcal{L}^d (\poseQ_k, \poseQ_{k+1}) +
D_2 \mathcal{L}^d (\poseQ_{k-1}, \poseQ_k)}_{
\mathcal{E} ( \poseQ_{k-1}, \poseQ_k, \poseQ_{k+1} )} = \mathbf{f}_{k+1}
\end{equation}
where $\mathbf{f}$ is the vector of net generalized forces applied to the system, and $D_i$ is a
partial derivative operator with respect to the $i$th function parameter.

{\bf Contact:} Contact is one of the greatest challenges, both computationally and
theoretically, with physical dynamics.  Such problems are less severe
if contact times and locations are known, or provided by a user (\eg, \cite{WeiChaiSIGGRAPH10}),
but in most real-word tracking problems contact is unknown at inference time.
Despite the added complexity, contact represents a strong constraint on motion (\eg, feet skate
should not happened during contact), and as such is a desirable element of the prior.
To avoid dependency on prior knowledge of the environment or manual intervention, we infer contact states
as part of a generative model. This reduces the computational challenge of handling inequality
constraints into enforcing holonomic constraints,
wherein one adds a constraint term,  $\mathcal{L}_c$, to the Lagrangian $\mathcal{L}^d$.
For a set of constraints, given by a function equation $\mathbf{g} ( \poseQ ) = 0$,
$\mathcal{L}_c$ is constructed as
\begin{equation}
\label{eq:multiple-holonomic-constraints}
\mathcal{L}_c ( \poseQ ) = \mathbf{g}(\poseQ)^T \bs\lambda
\end{equation}
where $\bs\lambda$ is a vector of Lagrange multipliers.

{\bf Root Forces:} We refer to the forces applied to the root node of the kinematic tree as root forces.
The root node represents the global translation/rotation, and forces applied to it represents external forces
applied to our physical model. Newton's 2nd law states that changes (over time) to the total momentum of a physical system
are equal to external forces applied to the system. Our model represents a system that has no external forces but contact.
As a result, we use the existence of root forces as an indicator that our contact model is incomplete (following \cite{Brubaker2009}).
By choosing a contact configuration that minimizes the external forces, we enhance our model with contact in order to enforce the assumption that contact
is the only option for our model to change its momentum, and propel itself.
Alternatively, applying a direct force to the root node of a kinematic tree can be
thought of as a human wearing a jet-pack. By minimizing the root forces, we discourage that option.

{\bf Contact Estimation:} \label{subsubsec:contact-estimation} To determine contact,
we trained an independent binary classifier (per possible contact point) on the forces of tracked subjects, assuming
contact-free motions.  Effectively, we learn to infer contact from the forces
that drive our model in the absence of contact.
Currently we use four possible contact points at the heel and toe of each foot.
A logistic regressor is trained to estimate the probability of contact
given theses forces:
\begin{equation}
\label{eq:predicting-contact}
p ( c_k^i | \mathbf{f}_k ) = \sigma_i(\mathbf{f}_k)
\end{equation}
where $c_k^i$ is a binary variable indicating the contact state of point $i$ in frame $k$,
$\mathbf{f}_k$ is the vector of net generalized forces at frame $k$ (in the absence of contact),
and $\sigma_i(\mathbf{x}) = (1+\exp (-\alpha_0^i-\mathbf{x}^T \bs\alpha_1^i))^{-1}$ is a
sigmoid function.
Parameters $\alpha_0^i, \bs\alpha_1^i$ were learned for each contact point
independently.

{\bf Physics-based Prior:}\label{subsubsec:physics-based-prior} With contact state determined,
we can estimate the contact forces by minimizing root forces (following \cite{Brubaker2009}).
We can then write Eq. \ref{eq:discrete-euler-lagrange-equations} with
the additional contact constraints as
\begin{equation}
\label{eq:euler-lagrange-with-contact}
\mathcal{E} \left( \poseQ_{k-1}, \poseQ_k, \poseQ_{k+1} \right) + \frac{\partial \mathbf{g}^T}{\partial
\poseQ_{k}} \bs\lambda = \mathbf{f}_{k+1}
\end{equation}
where $\bs\lambda$ is a vector of Lagrange multipliers for the holonomic contact constraints $\mathbf{g}$.
Given the selected contact configuration (\ie, active contact points), we can
estimate $\mathbf{f}$ such that the forces on the root node are minimized.
We achieve that by minimizing the squared norm of the root forces
\begin{equation}
\label{eq:minimizing-root-forces}
\left\Vert \mathbf{I}_{root} \left( \mathcal{E} + \frac{\partial \mathbf{g}^T}{\partial
\poseQ_{k}} \bs\lambda \right) \right\Vert^2_2
\end{equation}
where $\mathbf{I}_{root}$ is a square selection matrix in the size of
$\poseQ$, with ones on the diagonal to select the six degrees of freedom of the root node.
The regularized LS solution (by adding a small constant to the diagonal of $\mathbf{I}_{root}$)
yields the contact forces required to minimize root forces, \ie,
\begin{equation}
\label{eq:constraint-forces}
\bs\lambda^* = \left( \frac{\partial \mathbf{g}^T}{\partial \poseQ_{k}}
\mathbf{I}_{root} \frac{\partial \mathbf{g}}{\partial
\poseQ_{k}} \right)^{-1}
\frac{\partial \mathbf{g}^T}{\partial \poseQ_{k}} \mathbf{I}_{root}
\mathcal{E}
\end{equation}
We can then calculate the final forces
\begin{equation}
\label{eq:forces-estimation}
\mathbf{f}^*_{k+1} = \mathcal{E} + \frac{\partial \mathbf{g}^T}{\partial
\poseQ_{k}} \bs\lambda^*
\end{equation}
that are used as a prior in tracking. The prior over the forces $\mathbf{f}^*_{k+1}$ is defined twofold.
We would like to minimize the root forces as a generic prior for a plausible motion, and we would like
to minimize the internal torques to reduce jitter. Notice that given a contact configuration, both $\bs\lambda^*$
and $\mathbf{f}^*_{k+1}$ are functions of $\left( \poseQ_{k-1}, \poseQ_k, \poseQ_{k+1} \right)$, thus our physics-based
prior over the forces is applied directly over poses $\poseQ$.

\subsection{Registration and Tracking}
\label{subsec:registration-and-tracking}

Tracking is accomplished in an online fashion, by maximizing the posterior
distribution over state parameters at each frame.
As is common in online filtering, we assume conditional observation
independence, and a second-order Markov model to account for acceleration in the physics-based prior.
Accordingly, the posterior over state parameters at time $k$ is proportional
to the data likelihood and the conditional distribution over state parameters given those at previous time steps
\begin{equation}
\label{eq:tracking-probability}
\scalebox{0.9}{$p\left(\meshParams_{k}|\mathcal{D}_{k},\meshParams_{k-1:k-2} \right)
\, \propto \,
p\left(\mathcal{D}_{k}|\meshParams_{k}\right)\,
p\left(\meshParams_{k}\right|\meshParams_{k-1:k-2})
$
}
\end{equation}
where $\mathcal{D}_{k}$ is an input 3D point cloud at time $k$.
By assuming Gaussian noise in observations, the negative log likelihood of the data term in
Eq. \ref{eq:tracking-probability} becomes
\begin{equation}
\label{eq:data-probability}
- \log p \left( \mathcal{D}_{k} \, |\,  \meshParams_{k} \right) =
\sum_{(\meshVert', \dataPt') \in \Psi_k}{|| \meshVert' - \dataPt' ||^2_2}
\end{equation}
where $\Psi_k$ holds all matching body model points ${\meshVert' \in \meshVert( \meshParams_k )}$
and data points ${\dataPt' \in \mathcal{D}_{k}}$ at time $k$.
The matching was done in a standard Iterative Closest Point (ICP \cite{Besl92}) manner, and matched closest body and data points
with a maximal distance threshold and pruning of back facing vertices.
The data term captures the discrepancy between the model surface of
the body, encoded by mesh vertices $\meshVert_i( \meshParams_k )$,
and the observed depth data points.

The negative log likelihood of the conditional state probability is based on the physics-based priors,
as described above, and takes the following form:
\begin{equation}
\label{eq:physics-probability}
- \log p\left(\Theta_{k} \, |\, \Theta_{k-1:k-2} \right) =
\gamma_1  \Vert \mathbf{f}_{root}^{k}\Vert^{2} +
\gamma_2 \Vert \mathbf{f}_{-root}^{k}\Vert^{2}
\end{equation}
where $\gamma_1,\gamma_2$ are prior weights, $\mathbf{f}_{-root}^{k}$ comprises all but the
root forces and accounts for smoothness in torques,
and $\mathbf{f}_{root}^{k}$ are root forces which account to physical plausibility.

Tracking is formulated as the optimization of a global
objective $\mathcal{F}$, to find the parameters $\meshParams_k$ at each frame that minimize errors
between the body model, denoted $\meshModel( \meshParams_k )$, and an input 3D point cloud $\mathcal{D}_k$ at time $k$.
The objective is the negative log likelihood of Eq. \ref{eq:tracking-probability}, \ie,
\begin{equation}
\label{eq:model-ICP-error-function}
\scalebox{0.85}{$\mathcal{F}\left( \meshParams_{k-2:k}, \mathcal{D}_k \right) =
- \log p \left( \mathcal{D}_{k} \, |\,  \meshParams_{k} \right)
- \log p\left(\Theta_{k} \, |\, \Theta_{k-1:k-2} \right)
$
}
\end{equation}

as defined in Eq. \ref{eq:data-probability}, and Eq. \ref{eq:physics-probability} above.
A natural way to optimize this objective function is to use a variant
of ICP, \ie, by alternating between correspondence and
parameter optimization. Empirically, ICP tends to be both fast and accurate. We register our
model in the first frame by optimizing Eq. \ref{eq:model-ICP-error-function} w.r.t.
all parameters $( \poseQ, \anthroParams, \bshapeParams )$, and in the following frames update
the pose $\poseQ$ only, while holding $\anthroParams$ and $\bshapeParams$ fixed.

\subsection{Pre-processing}
\label{subsec:pre-processing}

The proposed ICP algorithm requires initialization of the body model parameters $\meshParams$.
In what follows we describe a fast and
simple initialization method. Following \cite{Baak2013}, we exploit the observation that the geodesic
distances between human end-effectors (\ie, head, hands, feet) are both large
and relatively independent of body pose, in order to automatically initialize tracking.

{\bf Geodesic Extrema as Scale/Rotation Invariant Mesh Features:} Given an
input point cloud $\mathcal{D}$, we generate a mesh
$\mathcal{D}_{mesh}$ (\ie, connecting vertices with edges) using a greedy
projection method for fast triangulation of unordered point clouds
\cite{Marton09ICRA}. In case of grid-based depth input data, we use a method
similar to \cite{Baak2013}, with a cut-off distance between nearby vertices
(\ie, threshold over maximal distance).

Given a connected component, we extract the first five geodesic
extrema $\{ \geoExtr_i \, |\,  \geoExtr_i \in \mathcal{D}_{mesh}  \}_{i = 1}^5$
from the geodesic centroid of the mesh, $\bar{\geoExtr}$, as in \cite{Baak2013}.
We order the geodesic extrema by geodesic distance, so that
$\geoDist(\geoExtr_i,\bar{\geoExtr}) \le \geoDist(\geoExtr_{i+1},\bar{\geoExtr})$,
where $\geoDist(\cdot,\cdot)$ is the geodesic distance between two points.
We define two features from which we detect human-like meshes, for labelling
end-effectors and for finding initial poses for tracking.
The first is a 4D vector that encodes the ratio of the ordered
geodesic distances:
\begin{equation}
\label{eq:ratio-feature}
\phi_{ratio} = ( r_1, ... , r_4 )^T ~~, ~~~~
r_j \equiv \frac{ \geoDist(\geoExtr_{j+1},\bar{\geoExtr}) }{ \geoDist(\geoExtr_j,\bar{\geoExtr}) }
\end{equation}
These features act like moments to describe the  geodesic eccentricity of the point cloud.
The second feature encodes geometric shape:
\begin{equation}
\label{eq:pose-feature}
\phi_{pos} = \left\{ \begin{array}{ll}
\text{angles between all triplets} \\
\text{angles between all orientations} \\
\end{array}
\right\}
\end{equation}
where orientation is defined as the vector between a geodesic extrema
and the point 30cm along the geodesic path to the geodesic centroid.

{\bf Detecting human-like components:} We use the distance-ratio features to
detect connected-components that might be people in the scene.  To that end we
learn a 4D Gaussian distribution over $\phi_{ratio}$ of human meshes. This
distribution then provides a probability that a connected component is a
plausible person. A threshold of $0.1$ on that probability is used to cull
non-human components.  Even with this simple method we accurately detect about 90\% of the
human components with minimal false positives, which is sufficient for our
application.

\begin{figure}[t]
\centering
\includegraphics[width=0.4\textwidth,height=3cm]{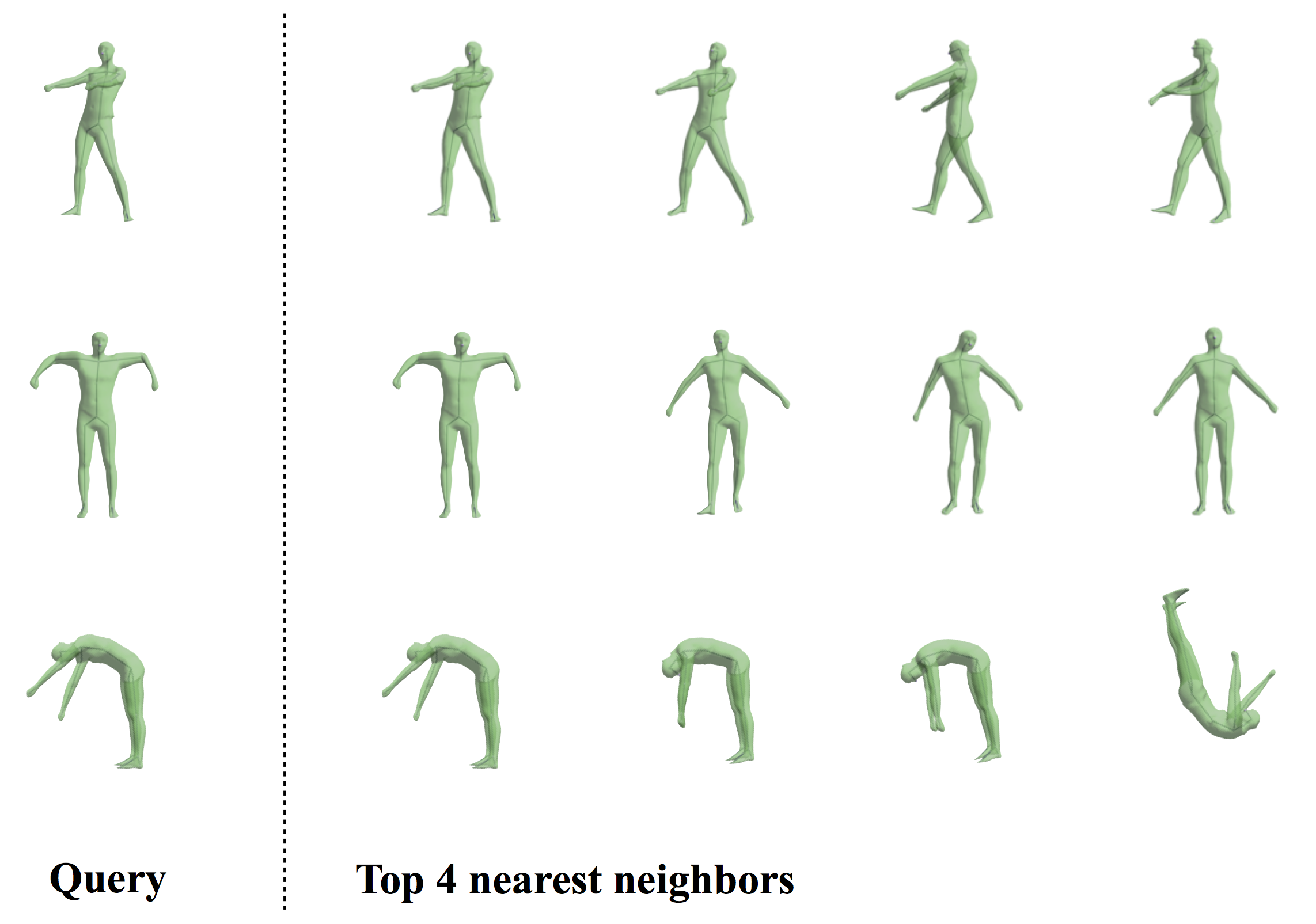}
\caption{Three examples of pose query based on $\phi_{pos}$. Since, by design,
the feature is scale and rotation invariant, we fetch different orientations of
similar poses.}
\label{fig:pose-query}
\end{figure}

{\bf Pose Initialization:} Given the extrema feature
descriptors, we can register an unregistered point cloud $\mathcal{D}$ by finding poses from
a database of labelled point clouds whose features $\phi_{pos}$
are most similar to those of the point cloud (see examples in Fig. \ref{fig:pose-query}).
In more details: we fetch a pose from a pose data-base (based on $\phi_{pos}$ L2 distance),
we align the database mesh (with mean $\meshParams$) and the data point-cloud (based on
fitted Ellipsoids to vertices), we estimate $\meshParams$ (ICP).

\subsection{Method Summary}

To summarize, the tracking pipeline is as follow:
\begin{algorithmic}[1]
\STATE Divide $\mathcal{D}$ into connected components
\STATE Remove non-human connected components  by using $\phi_{ratio}$
\STATE Initialize pose and register first frame (ICP)
\FORALL{frames}
    \STATE Initialize pose with previous pose
    \STATE Execute ICP
\ENDFOR
\end{algorithmic}

\section{Experiments}
\label{sec:experiments}

\subsection{Execution Speed}
\label{subsec:execution-speed}

\begin{table}
\centering
{\small
\begin{tabular}{ | l || l | l |}
\hline
\bf{Contact Point} & \bf{No Kalman Filter} & \bf{With Kalman Filter} \\
\hline \hline
Left Toe & 94.6  & 98.9  \\
Right Toe & 95.6  & 98.5 \\
Left Heel & 77.4 &  84.0  \\
Right Heel & 87.9  & 95.9 \\
\hline
\end{tabular}
}
\caption{A comparison between contact predictor performance, with and without
Kalman filter. The values represent the percentage of predicting ground truth
contact state.}
\label{tab:contact-predictor-accuracy}
\end{table}

The tracking system was implemented in Python, with the core physical
components in C++.  It was evaluated on a desktop running OS X, with
Intel Core $i7/2.3 GHz$ and 8 GB RAM.  At present it runs at
$0.2 [fps]$ with physical priors, and $1.96 [fps]$ with body model only.

\subsection{Quantitative Comparison}
\label{subsec:quantitative-comparison}

We used {SMMC-10} dataset \cite{SMMC10} for quantitative comparison.
It comprises synchronized Vicon mocap marker data and Mesa SwissRanger ToF
depth data. The depth data have significant amounts of noise, as
can be seen in Fig. \ref{fig:registration-results-SMMC10}, and the accompanying video. We compare our method
to \cite{Baak2013} and \cite{Ganapathi2012} on the same dataset.
We achieve state-of-the-art tracking accuracy (see Table \ref{tab:accuracy-comparison}).
We show the results of two metrics:
Mean Joint Prediction Error (MJPE) - the RMSE of predicting the mocap markers from our skeleton joints, and
Mean Joint Indicator Error (MJIE) - the average of all  joint predictions that were within a range of $10 [cm]$ from the target joint.
Interestingly, when based solely on MSE metrics (as defined above), the physics-based prior does not
appear to significantly affect performance (\eg, see  Table \ref{tab:accuracy-comparison}). On the contrary,
the accompanying video demonstrates how MSE metric does not reflect the physical plausibility of a motion.
That is, there are many motions for a given MSE metric, most of which are not physically plausible per se.

Due to high SNR in {SMMC-10} dataset's depth scans, and as a result in our force estimation, we used an online Kalman filter as a
noise filtering technique. We have found that this improves performance, reducing prediction
error, by roughly 50\% (see Table \ref{tab:contact-predictor-accuracy}).

We used a threshold of $0.8$ contact probability to reduce sticky
contact (\ie, false positive contact prediction). This is the result of enforcing holonomic (equality)
contact constraints instead of inequality constraints. Despite simplifying the
joint distribution over all contact points into an independent
probabilistic model per contact point, our model was accurate
enough to allow tracking with 4 possible contact points. The contact prediction model predicted
the correct contact configuration with more than $95\%$ mean accuracy (see Table \ref{tab:contact-predictor-accuracy}).

\begin{table}
\centering
{\small
\begin{tabular}{ | l || l | l | }
\hline
& \bf{MJPE} & \bf{MJIE} \\
\hline \hline
\bf{Proposed Method (Data Only)} & $1.7 [cm] $ & 0.975 \\
\bf{Proposed Method (Physics)}  & $1.89 [cm] $ &  0.973  \\
\bf{Ganapathi} & N/A & 0.971 \\
\bf{Baak} & $ \thicksim 5[cm] $ & N/A \\
\hline
\end{tabular}
}
\caption{Quantitative results. The metrics MJPE and MJIE are defined in Sec. \ref{subsec:quantitative-comparison}.
While accurate numbers for Baak were not available, it is at best $ 5 [cm]$, as shown in his work from 2013. }
\label{tab:accuracy-comparison}
\end{table}

\subsection{Qualitative Comparison}

\begin{figure}[t]
\centering
\includegraphics[width=0.3\textwidth,height=3cm]{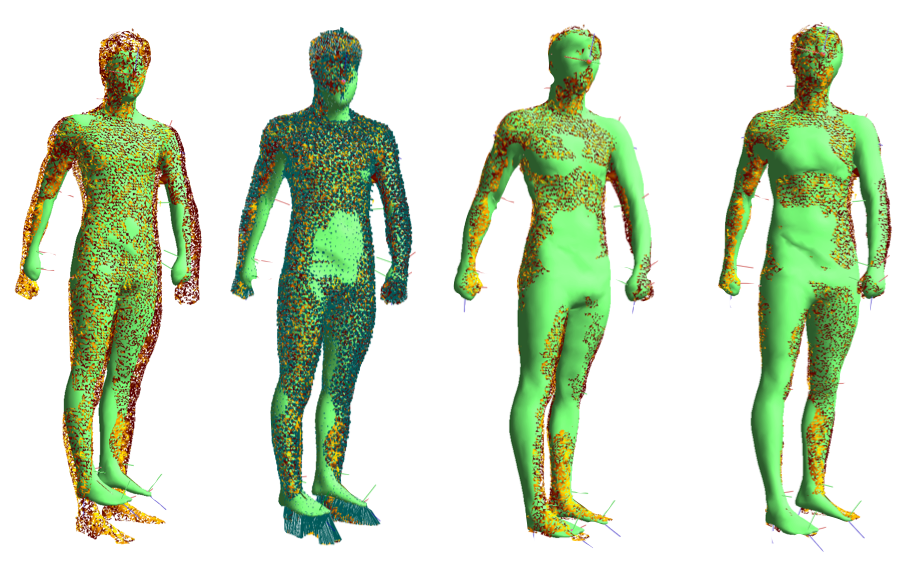}
\caption{Registration results of an accurate laser scan. From left to right:
initial alignment, ICP correspondences, gradient descent step, and final
registration. }
\label{fig:registration-results-scan}
\end{figure}

\begin{figure}[t]
\centering
\includegraphics[width=0.4\textwidth,height=3cm]{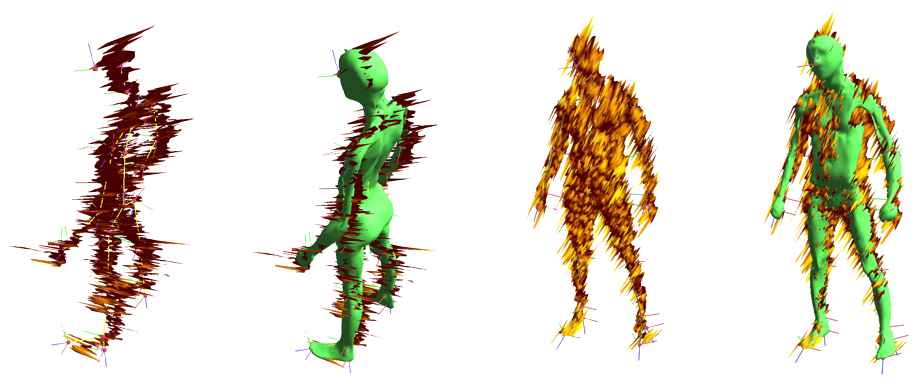}
\caption{Registration results of noisy depth sensor ({SMMC-10} dataset). In yellow
is the noisy depth input data. Notice that the amplitude of noise is larger than
the thickness of arms and legs. Despite that, we get a reasonable registration,
when visually inspected.}
\label{fig:registration-results-SMMC10}
\end{figure}

Fig. \ref{fig:registration-results-scan} depicts the process of registering
an accurate laser scan with our model. Notice how we learn anthropometrics,
body shape and pose. Similarly, in Fig.\ \ref{fig:registration-results-SMMC10}
we register our model to the first frame in a tracking sequence. Due to the low
dimensionality of the model we are still able to register a plausible model to very noisy data.

\begin{figure}[t] \centering
\includegraphics[width=0.5\textwidth]{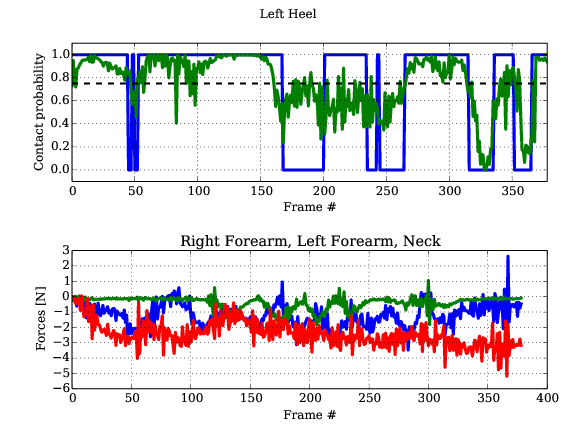}
\caption{An example of contact prediction. In the top graph, blue is the ground
truth contact, while green is the estimated probability. The bottom graph
depicts the three body parts with the largest weights in predicting contact.}
\label{fig:contact-prediction}
\end{figure}

We tested our registration technique (Sec.
\ref{subsec:registration-and-tracking}) on Hasler \cite{Hasler2009}, and {SMMC-10} \cite{Ganapathi2012} datasets
with promising initial results.
When inspecting the tracking results in the accompanying video, there are some visible artifacts in the mesh model.
Those, however, are mainly due to different pose distributions in
training the mesh model and during tracking, rather than due to fundamental
limitation of the model (excluding known artifacts of linear mesh blending
such as  volume collapse). Another interesting property of our mesh model is
volume prediction per registered mesh. We used the mesh volume to
calculate the inertial description needed for the physical priors, treating the volume as water.
When compared with ground truth, we had an average weight prediction error of
$1.5\%$ of $115$ subjects.

The true power of our approach is with the reduction of visual artifacts.
While we do not remove jitter entirely, it
is attenuated, when compared with data-only tracking. A more dramatic
result is how foot-slide is removed in cases where contact is correctly
detected. Despite the fact that our false positive contact estimation
(wrong contact prediction) caused occasional visual artifacts, the value
of removing foot-slide is much more noticeable, as evident in the
accompanying video.
To better understand how the force predictor works,
Fig.\ \ref{fig:contact-prediction} plots ground truth contacts, along
with the corresponding forces. We considered the lowest marker, along with all markers up to
$5 [cm]$ away as in contact, due to lack of contact ground truth.
Despite having noisy ground truth, our simple predictor was able to perform well on most frames.

\begin{figure}[t] \centering
\includegraphics[width=0.5\textwidth]{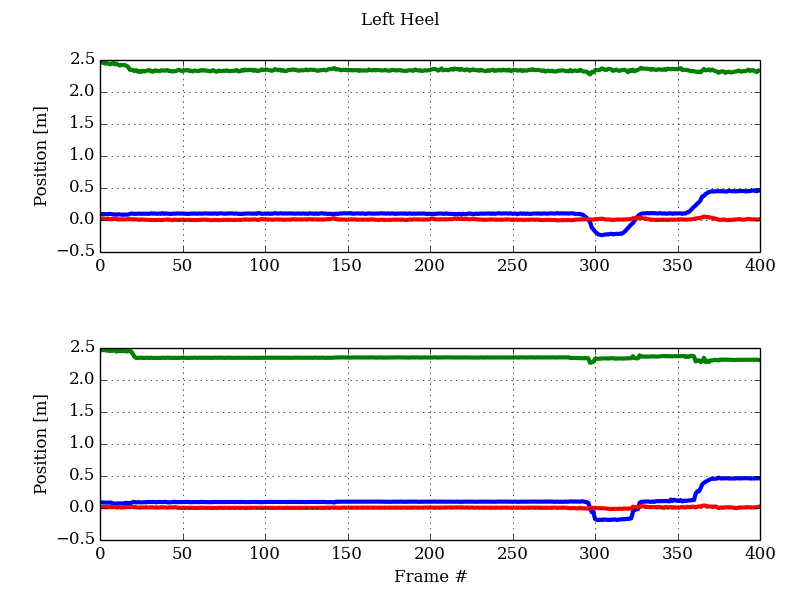}
\caption{An example of the effects of contact on the foot. Top plot: RGB for
XYZ of left heel position, with no contact constraints. Bottom: with contact
constraints. Notice how the contact reduced jitter and foot slide, while still
preserving the global pattern.}
\label{fig:contact-position}
\end{figure}

Fig. \ref{fig:contact-position} demonstrates how adding the contact constraint acts as a strong motion prior.
While smoothing pose will reduce jitter, it will also reduce discontinuities in motion due to contact. On the other hand,
applying physical contact constraints can smooth jitter while allowing abrupt changes in motion.

\section{Discussion}
\label{sec:discussion}

We propose an online physics-based 3D human tracking approach that
incorporates physics-based priors into tracking without the need for subject calibration
or knowledge about the environment. The use of physics in this context
is compelling as it allows us to minimize visual artifacts, most notably
jitter and foot-skate which results from noise and occlusions.
We demonstrate that we can infer contact from joint
torque trajectories computed by inverse dynamics.
We show that our method is effective at tracking from a noisy single depth
sensor and produces quantitative results that are on par or better than
current state-of-the-art, while at the same time qualitatively reducing visual
artifacts.

Our contact prediction model, while conceptually compelling, is relatively
simple. For example, we predict all of the contact points independently,
despite the fact that contact patterns (especially for contact points on
connected segments) are clearly correlated. We believe the prediction model
can be further improved by structured prediction that incorporate these correlations
in contact state.
While our method predicts the environment online, it currently does not
aggregate these predictions, which may be important for longer sequences.

We also note that in our method the ground can both push and pull on
the body when contact is established. This can sometimes be seen in
the video. While this behaviour is not realistic in terms of underlying physical behaviour,
we nevertheless believe it allows us to overcome vast amounts of noise
in the observations (especially where feet can easily be confused
with occluders and the ground plane).

Finally, we note how MSE metrics do not capture the dynamics of a motion.
For example, the same MJPE can represent a motion that is the ground truth with a constant added to it,
or the ground truth with a Gaussian noise with the same constant as a standard deviation, and a zero mean.
This exposes the limitation of relying on a MSE metric to assess the quality of tracking results,
and highlight the advantage of using physics-based prior. In a sense, the physics-based prior
shapes the results to be qualitatively superior, despite not improving the MSE metric itself.

\bibliographystyle{IEEEtran}
\bibliography{IEEEabrv,paper_strip}

\newpage

\section{Learning the Body Model}
\label{sec:learning}

\begin{table}
\centering
\scalebox{0.85}{%
\begin{tabular}{ | l || l | l | l | l | }
\hline
& \bf{Our Model} & \bf{SCAPE} & \bf{Hasler} & \bf{Allen} \\
\hline \hline
\bf{Mesh}  & \multirow{2}{*}{Linear}  & Nonlinear  & Nonlinear & Linear \\
\bf{recon.}  &  & (Least Squares)  & (Poisson) & \\
\hline
\bf{Mesh} & 1.82 [mSec] (full)  & 1 [Sec]  &  & \\
\bf{recon.} & 1.17 [mSec] (pose) & mesh only & 25 [Sec] & 13 [mSec] \\
\bf{speed} & & (given matrices) & & \\
\hline
\multirow{3}{*}{\bf{Semantics}} & Skeleton  & Skeleton & & \\
& Body shape & Body shape & None &  Skeleton \\
& Anthropometrics & & & \\
\hline
Mean           &  &  & & \\
reconstruction & 5.3 [mm]  & N/A  & 54 [mm] & 4.9 [mm] \\
error          &  &  & & \\
\hline
\end{tabular}%
}
\caption{
A comparison between our model, SCAPE \protect\cite{Anguelov2005},
Hastler \protect\cite{Hasler2009}, and Allen \protect\cite{Allen2006}.
\textbf{Semantics} refers to the direct interpretation of model parameters.
Semantic parameters, such as explicit anthropometrics (skeleton) representation,
is useful in the context of tracking. In our case, it allows direct
control over which set of parameters to optimize.}
\label{tab:mesh-model-comparison}
\end{table}

In order to learn the body mesh model in Sec. \ref{subsec:fast data-driven parametric body model}
we use the Hastler dataset \cite{Hasler2009}, which consists
of 111 subjects with 520 poses, all with registered meshes. We learn the model
by minimizing Eq. \ref{eq:model-ICP-error-function} w.r.t. the weights
$W = \left\{ w_{ib} \right\}$, the different mesh templates $\left\{ \tilde{\meshVert}^s
\right\}_{s}$ per subject $s$, and the template pose and anthropometrics $\left\{ \poseQ^s,
\anthroParams^s \right\}_{s}$ per subject $s$. We define the number of weights per
vertex based on joints proximity along the kinematic tree, with BFS of distance 3. Since we optimize all
parameters w.r.t. to the same reconstruction error function (Eq.
\ref{eq:model-ICP-error-function}), we get an accurate reconstruction despite
the simplicity of the model, when compared with other state-of-the-art models
(Table \ref{tab:mesh-model-comparison}).

Notice that some of the models in Table
\ref{tab:mesh-model-comparison} were trained on different datasets.
However, our main goal is to demonstrate that our model is comparable to state-of-the-art models,
rather than a comprehensive comparison.  In our dataset only 43 out of
111 subjects have more than a single pose, which is required for our training.
However, those subjects account for 86\%
of the total number of poses (450 out of 520 poses).

\subsection{Model Parameters Optimization}

While it is possible to
optimize for the weights, mesh templates and pose simultaneously, it is a slow,
non-convex and nonlinear optimization.
Instead we alternated the optimization between the parameters, which yields a much faster
optimization process, and is also convex w.r.t. the mesh templates and
weights. Our learning process includes the following steps:
\begin{enumerate}
  \item Initialize $\poseQ^s$ for all subjects by fitting landmarks (based
  on the registered meshes).
  \item Repeat until convergence:
  \begin{enumerate}
    \item Optimize weights $W = \{w_{ib}\}$ given
    current poses and mesh templates. We optimize a global reconstruction
    error function as $W$ is shared among all subjects
    \begin{equation}
\label{eq:global-model-ICP-error-function}
\mathcal{F} = \sum_{s \in \mathcal{S}}{\mathcal{F}^s\left( \meshParams^s, \mathcal{D}
\right)}
\end{equation}
where $\mathcal{S}$ are all subjects with more than a single pose in our
dataset. By examining Eq. \ref{eq:linear-mesh-blending}, it is clear that Eq.
\ref{eq:global-model-ICP-error-function} is convex w.r.t $W$. Thus, we can define $\mathbf{A}_{i}^{s,j}$ by rewriting Eq.
\ref{eq:linear-mesh-blending} as

\begin{equation}
\label{eq:weights-optimization}
\scalebox{0.8}{$%
\begin{array}{l}
\left(\begin{array}{ccc}
\hdots & \mathbf{M}_{b}\left(\poseQ_{j}\right)\cdot\tilde{\mathbf{M}}_{b}
\left(\tilde{\poseQ}^{s}\right)^{-1}\tilde{\meshVert}_{i}^{s} &
\hdots\end{array}\right)\mathbf{w}_{i}=\meshVert_{i}^{j}\\
\Rightarrow\mathbf{A}_{i}^{s,j}\cdot\mathbf{w}_{i}=\meshVert_{i}^{j}
\end{array}
$
}
\end{equation}
where $j$ is a pose index (over all poses of
subject $s$), and $\mathbf{w}_i$ is the weights of vertex $i$ as a vector. By
concatenating $\mathbf{A}^j_i, \meshVert^j_i$ of all subjects $s$ and all poses
$j$ it is easy to calculate the least-squares solution.

\item Optimize mesh template per subject, given current weights and poses. We
optimize the mesh template independently per subject. By examining
Eq. \ref{eq:linear-mesh-blending}, it is clear that Eq.
\ref{eq:model-ICP-error-function} is convex w.r.t to $\tilde{\meshVert}^s$, and we
can define $\mathbf{T}_{i}^{s,j}$ by rewriting
Eq. \ref{eq:linear-mesh-blending} as

\begin{equation}
\label{eq:template-optimization}
\mathbf{T}_{i}^{s,j} \cdot \tilde{\meshVert}_{i}^{s}=p_{i}^{j}
\end{equation}
per vertex $i$ and per pose $j$. By concatenating matrix $\mathbf{T}_i^{s,j}$
for all poses $j$ per subject,  a simple least-squares solution can be used here
as well.

\item Optimize pose $\poseQ^j$ for all poses of all subjects. All poses can
be estimated independently, by using nonlinear and non-convex optimization of Eq.
\ref{eq:model-ICP-error-function} w.r.t. the pose parameter $\poseQ$. We
used BFGS to optimize for pose parameters $\poseQ_s^k$ for all poses of all
subject.
Note:
Since the optimization is local, good initialization is required.
  \end{enumerate}
\end{enumerate}

Practically, two full iterations iterations of $(a),(b),(c)$ were enough to get
close to convergence. The result of the model parameters optimization phase are $W$, shared weights to be used in LMB, the
mesh template per subject $\tilde{\meshVert}^s$, the bones length $\anthroParams^s$ per
subject, and the pose vector $\poseQ^j_s$ per pose $j$ and subject $s$.

\subsection{Basis Learning}

Once we learn the model parameters as explained above,
we can train a linear regressor with basis $\bonesMat$ from bones length to
a mesh template $\tilde{\meshVert}$, s.t.

\begin{equation}
\label{eq:bone-to-template-linear-regression}
\tilde{\meshVert}^s \approx \bonesMat \cdot \anthroParams^s
\end{equation}
where $\bonesMat$ is learned with a least squares formulation.

By applying PCA to the null space of the linear regression basis (difference
between regressed mesh template and $\tilde{\meshVert}$), we can learn the body
shape basis $\bodyMat$. We used the first 10 PC as a linear basis. Once
we have the two basis, $\bonesMat, \bodyMat$, we can generate new
mesh templates given any desired bones length $\anthroParams$ and body shape score
$\bshapeParams$, as shown in Eq.
\ref{eq:model-template}, and generate a mesh for any given pose by using Eq.
\ref{eq:linear-mesh-blending}.

\end{document}